\documentclass[]{spie}  
\usepackage[]{graphicx}
\usepackage{amsfonts}
\usepackage{amsmath}
\usepackage{algorithm}
\usepackage{algorithmic}
\usepackage{subfigure}

\title{How to find real-world applications for compressive sensing} 

\author{Leslie N. Smith 
\skiplinehalf
U.S. Naval Research Laboratory, Washington, D.C.  20375 USA \\
}

\authorinfo{Further author information: 
E-mail: Leslie.Smith@nrl.navy.mil, Telephone: 1 (202) 767-9532  
}

\begin{document} 
  \maketitle 

\begin{abstract}
The potential of compressive sensing (CS) has spurred great interest in the research community and is a fast growing area of research. However, research translating CS theory into practical hardware and demonstrating clear and significant benefits with this hardware over current, conventional imaging techniques has been limited. This article helps researchers to find those niche applications where the CS approach provides substantial gain over conventional approaches by articulating lessons learned in finding one such application; sea skimming missile detection. As a proof of concept, it is demonstrated that a simplified CS missile detection architecture and algorithm provides comparable results to the conventional imaging approach but using a smaller focal plane array. The primary message is that all of the excitement surrounding CS is necessary and appropriate for encouraging our creativity but we all must also take off our "rose colored glasses" and critically judge our ideas, methods and results relative to conventional imaging approaches.
\end{abstract}


\keywords{Compressed sensing, target detection}

\section{INTRODUCTION}
\label{sec:intro}  

The data deluge of the past few decades sparked a great deal of research in image compression.
In image compression, such as JPEG, most of the acquired data is accurately approximated by much fewer coefficients of a basis such as wavelets.
An interesting research question was posed by Donoho\cite{Donoho06} ``why go to so much effort to acquire all the data when most of what we get will be thrown away? Can we not just directly measure the part that will not end up being thrown away?''
Compressive sensing (CS) bypasses this wasteful acquisition process but rather only measures the relatively fewer values necessary to reconstruct the signal or image.
This amazing ability has been the topic of a fast growing area of research as is illustrated in a paper by Elad\cite{Elad12} (see his Figure 2).

However, translating the fundamentals of CS into practical hardware with significant benefits over conventional imaging has been limited.  
As stated in a JASON report for 2012\cite{JASON12}, ``The CS literature often deals with idealized situations".
Despite thousands of paper published yearly on this subject, few head to head comparisons between CS and conventional imaging techniques appear  in the literature.
The advantages of CS are derived from a savings in terms of size, weight, power, and costs compared to a conventional sensor.  
The disadvantages come from the need to post-process the measurements and a reduction in peak signal to noise ratio (PSNR).
This tradeoff indicates situations where implementing a CS architecture might be beneficial.

For example, our efforts to directly compare CS image construction and super-resolution to conventional imaging with super-resolution for aerial imagery proved conventional imaging methods to be preferable.
More specifically, we compared using the same low resolution focal plane array (FPA) in a CS camera mode and in a conventional sensor with the goal of attaining a 4 times greater resolution image reconstruction (double the width and height resolution of the FPA).
For the conventional camera approach we simply used bicubic interpolation to obtain the higher resolution image.
Interpolation is computationally fast and only created some blurring in the final result.
On the other hand, a variety of compressive sensing approaches in the literature were attempted, such as the CS imager described by Willett, et.al. \cite{Willett11} and the algorithm provided by Romberg \cite{Romberg08}.
In some cases the results contained artifacts and in all cases the CS approach was more computationally expensive.
These experiments indicated that whenever conventional techniques are easily implemented, conventional imaging techniques are preferable to CS techniques.
These tests indicate that CS is most valuable in order to overcome limitations of conventional techniques; that is, when conventional imaging is hard to do.

The aim of this article is to encourage and assist researchers in finding those niche applications where the CS approach provide substantial gain compared to conventional approaches.
The main contributions of this paper are twofold: first, to present lessons learned to help find practical, real-world applications derived from the author's recent experience in evaluating CS applications; 
and second, these guidelines are illustrated by showing the benefits of CS versus conventional imaging of a new application - sea skimming missile detection. 

Some of the past successful applications for CS provides guidance in finding other successful applications.
Two such early wins are SparseMRI\cite{Lustig07} and radio interferometry\cite{JASON12}.
Analysis of the methods behind these two very different applications show many similarities.
From this analysis the following guidelines for successful application of CS can be derived: 
\begin{enumerate}  
\setlength{\itemsep}{1pt}
  \setlength{\parskip}{0pt}
  \setlength{\parsep}{0pt}
  \item CS most useful when it can overcome hardware, financial, bandwidth, or battlefield limitations,
  \item A priori knowledge of what you will be imaging,
  \item The signal is sparse in the pixel domain or there exist a transformation to create a sparse representation and additionally the sparsity is orders of magnitude smaller than the number of pixels, 
  \item The measurement matrix is (optimally) incoherent with the sparsifying transform,
  \item There exists physically realizable hardware architecture corresponding to the theoretical design.
\end{enumerate}
The first two conditions come out of practical considerations and eliminate many potential applications for CS.
This implies that it is necessary to find niche applications where CS has unusual strengths such that it is better to use CS for those strengths than use a conventional sensor.
The next two conditions derive from CS theory and are prevalent throughout the literature.
The last condition grounds the architecture in a realizable CS camera.

In the next section we will provide a brief description of compressive sensing prior to describing a new, practical application for CS.

\subsection{Compressive Sensing} 

Compressive sensing (CS) grew out of the work of initial work by Emmanuel J. Cand\`{e}s, Justin Romberg and Terence Tao\cite{Candes06}, and David Donoho\cite{Donoho06}.
At it's heart, compressive sensing is about breaking the Nyquist barrier; that is the ability to image when collecting much fewer measurements than twice the highest frequency dictated by the Nyquist theorem.
This is possible because there is much less useful information than indicated by the frequency.
The key to compressive sensing's success relies on the signal being sparse or compressible in either its natural state or sparse in a known basis.
In this situation, solving an underdetermined, linear system of equations can uncover the essential information within the few measurements.
Transform coding\cite{Eldar12, Elad10} relies on finding a basis where the signal is sparse or compressible.

Here we provide a description of key terms:
\begin{enumerate}  
\setlength{\itemsep}{1pt}
  \setlength{\parskip}{0pt}
  \setlength{\parsep}{0pt}
  \item Sparsity - a signal is k sparse if it has at most k terms or if it can be exactly represented by at most k atoms from a transform or dictionary.
  \item Compressible - the coefficients of a representation of a signal decrease exponentially, implying that the signal can be accurately represented by it largest k coefficients.
  \item Coherence/incoherence - the metric used for a measurement transform that is the largest absolute product between the a sparse or compressible signal and the measurement matrix.  Intuitively, a small coherence means the signals are spread out in the measurement domain, which leave "clues" to their existence.
Fourier transforms is a good example of incoherence since a Dirac function or spike in the signal domain is spread out throughout the Fourier/frequency domain.
  \item Partial Fourier (PF) coefficients - undersampling the Fourier coefficients and retaining only a fraction of the coefficients.
\end{enumerate}

More formally, 
one seeks to recover a $k$-sparse signal $x \in \mathbf{R}^N $ based on the measurements $y  \in \mathbf{R}^M$, each of which is a different linear combination of the $x$.
Here one expects $ k < M \ll N  $, where a lower bound of $ M \approx k \times log ( N ) $ is given in one of the original papers by Cand\`es, Romberg, and Tao\cite{Candes06} but more recent research claim even fewer measurements are sufficient.
In the event 
that $x$ is not naturally sparse, we assume the existence of a linear 
transform $\Psi$ such that $x=\Psi\theta$ and  $\theta\in \mathbf{R}^N$ is $k$-sparse in the coefficient vector.  
Given this notation, assume the measurement 
process is described by 
the equation
\begin{equation}
y = A \Psi  \theta + n
\end{equation}
where the $M\times N$ matrix $A$ models measurement process, and $n\in \mathbf{R}^M$ represents noise in the measurements.  CS theory has shown that it is possible to accurately reconstruct $\theta$ given $y$ by solving the optimization problem \cite{Candes06}
\begin{equation}
\label{eqn:CS}
\operatorname*{arg\,min}_{\theta} \parallel \theta \parallel_0  ~~subject~to  \parallel y -  A \Psi  \theta  \parallel_2^2 < \epsilon
\end{equation}
where $\epsilon$ is a small scalar related to the noise variance.
If there isn't any noise in the measurements ($ n = 0 $) this situation is called the exact model and equation \ref{eqn:CS} reduces to
\[
\operatorname*{arg\,min}_{\theta} \parallel \theta \parallel_0  ~~subject~to ~~ y =  A \Psi  \theta
\]

In the past few years a significant amount of research was directed to solving Equation \ref{eqn:CS} and reconstructing the original image.
It is NP hard to solve this equation and it is typically solved approximately either by greedy methods or $L_1$ relaxation methods.
Greedy methods include Orthogonal Matching Pursuits (OMP)\cite{Mallat93, Tropp07}, CoSaMP\cite{Needell09}, Subspace Pursuits\cite{Dai09}, thresholding\cite{Elad10}, and others.
Since the objective function $ \parallel . \parallel_0 $  in Equation \ref{eqn:CS} in non-convex, it is common in the literature to ``relax" the objective 
function with a convex approximation $ \parallel . \parallel_1. $ Hence one instead solves
\begin{equation}
\label{eqn:L1}
\operatorname*{arg\,min}_{\theta} \parallel \theta \parallel_1  ~~subject~to  \parallel y -  A \Psi  \theta  \parallel_2^2 < \epsilon  
\end{equation}
which can be solved by linear programming methods\cite{Chen98} for which efficient solvers exist.
An equivalent solution can be obtained by solving a modified form of Equation \ref{eqn:L1}  
\begin{equation}
\label{eqn:lambda}
\operatorname*{arg\,min}_{\theta} \left( \frac{1}{2} \parallel y -  A \Psi  \theta  \parallel_2^2 + \lambda  \parallel \theta \parallel_1 \right)
\end{equation}
where $ \lambda $ is a parameter to balance the solution between the first term for integrity to the data and the second term to promote sparsity.

In addition to the sparsity requirement, the above solutions require that the sensing matrix $A$ be ``incoherent'' with the signal model $\Psi$.  
Incoherence can be understood heuristically in the following way;
the incoherence property suggests that the measurement system $A$ takes a sparse signal $\theta$ and spreads it out over many observations $y$.  
If this property holds, information sufficient to recover the vector $\theta$ are contained in the $M$ measurements $y$.  
For example, if the sparse signal consists of $k$ spikes we might choose $\Psi={\bf I}$ (the identity matrix).  
In this case the Fourier transform matrix $A$ is optimally incoherent and will spread out information of the spikes throughout the frequency domain.  
One could also design a CS imaging system so that $A$ is comprised of random entries (e.g., each entry equiprobable ``0'' or ``1'').  This essentially guarantees that a vector that is sparse in the basis $\Psi$ will be non-sparse in $A$.

\subsection{Related Work} 

A great many papers in the compressive sensing literature discuss how to implement practical compressed sensing systems (e.g., see \cite{Willett11, Romberg09, Wagadarikar08}) but few  show CS imaging to be better than conventional imaging when all aspects of the two approaches are weighed and compared.
Most practical CS hardware architectures can be found described on Igor Carron's website at https://sites.google.com/site/igorcarron2/compressedsensinghardware
and the latest information within the CS field may be discovered in his daily blog at http://nuit-blanche.blogspot.com/.

As discussed above, two credible applications for CS are in its use for MRI\cite{Lustig07} and radio interferometry\cite{Li11}.
We used these wins as prototypes for establishing the guidelines for successful application of CS.
There are a number of papers in the literature applying CS to detection of stars, such as the paper by Gupta, et. al.\cite{Gupta11}.
Missile detection considered later in this paper bears similarities to star point source detections algorithms.
This also bears some resemblance to the application of compressive sampling to radio interferometry\cite{Li11} where the use of partial Fourier reconstruction is described.

The JASON report on "Compressive sensing for DoD sensor systems"\cite{JASON12} is closely related to the topic of this article as the report addresses the question where CS can be beneficially utilized for DoD radar applications.  
Some of the guidelines for successful application of CS can be found in this report but are not clearly delineated as in this paper.
Ultimately the report finds that CS should be of interest to DoD but more research is required, such as this paper, to determine useful applications.

The paper ``Compressive sampling vs. conventional imaging'' by Haupt and Nowark\cite{Haupt06} addresses the same question that is the focus of this paper.
However, that paper does not provide general guidelines nor an example of a new CS application as is done in this paper.
Several recent papers, such as one by Adcock, et. al.\cite{Adcock13}, point out the gap between CS theory and use in real world applications and show how to bridge it.

There are in the CS literature papers related to using compressive sampling for target detection and tracking.
The paper by Kashter, Levi, and Stern\cite{Kashter12} describe application of CS to motion detection. 
However, their method relies on frame differencing to achieve sparsity; that is, only the moving objects appear in the frame differenced image.
Similarly the paper by Poon, et. al.\cite{Poon12} describes a realizable hardware application that they applied to a simplified target tracking problem.
Again their method depends on frame differencing, hence the objects moving to different pixels from frame to frame.
In the missile detection problem, the sub-pixel missile signal remains in the same pixel for tens of frames and frame differencing eliminates all possibility of detecting the already dim target.
A paper by Duarte, et. al.\cite{Duarte06} describes the signal detection problem using CS principles without reconstructing the image.
The missile detection application described here similarly follows this concept; that is, the goal is to detect the missile regardless of the quality of the image reconstruction.

\section{A NEW APPLICATION OF COMPRESSIVE SENSING} 

In this section we describe the conventional and new CS approaches to the missile detection problem.  
Specifically we are able to demonstrate better Receiver Operating Characteristic (ROC) performance with a lower resolution focal plane array than a previously used, conventional approach.  
Rather than try and recover a pristine (low PSNR) image, we show that by modifying the CS recovery algorithm we can effect greater probability of detection for low false alarm rates.  
This is, of course, the main goal of a missile warning system and hence is a clear demonstration of the potential of CS architectures.

Section \ref{sec:missile} briefly describes the sensor system and field data collected to support the system development. Section \ref{sec:CSmissile} 
describes how to adapt the sensor system and algorithms for a CS approach.

\begin{figure}[tbc]
  \centerline{
   \begin{tabular}{c}
        \includegraphics[scale=0.30]{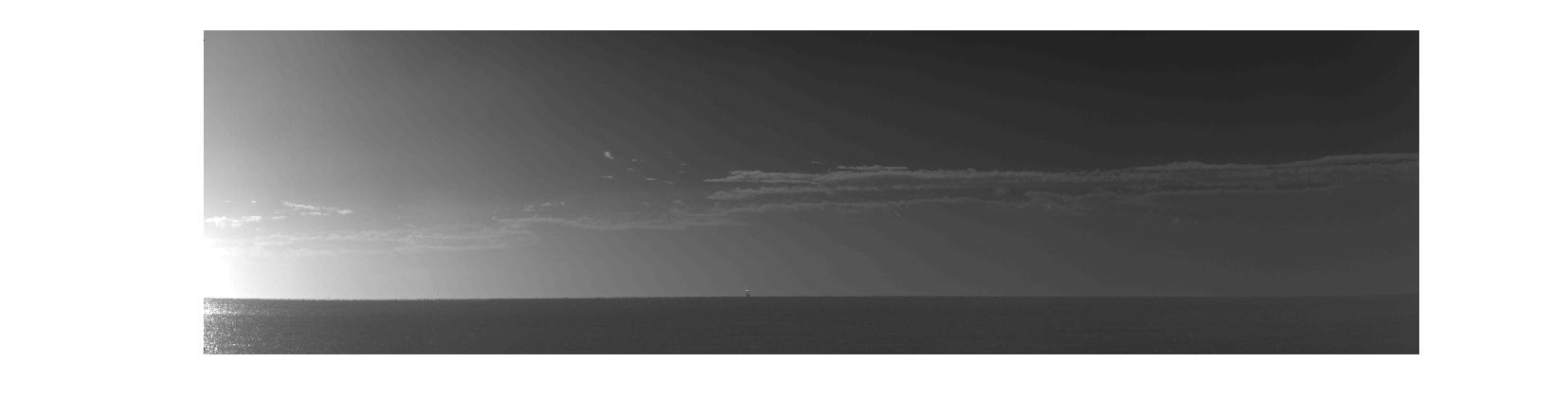}
   \end{tabular}
   }
   \caption{Full 45x 3.7 degree sensor image at 0800.}
   \label{fig:OrigImage}
\end{figure}

\subsection{Missile Detection} 
\label{sec:missile}

Several years ago the Naval Research Laboratory (NRL) developed a MWIR sensor system for detection, tracking, and declaration of sub- and supersonic 
sea-skimming, anti-ship cruise missiles.  
The objective was to find bright, sub-pixel sized objects just above the horizon.  As part of this effort, both the sensor and detection algorithms were 
developed, and subsequently tested with field data.  This section describes the sensor, algorithms, and the collected field data.

The sensor used in this study is a passive staring midwave IR sensor \cite{LNS06}.   
It consists of an f/2 80 mm diameter optical system with a prism anamorphic element that results in a 3.6 x 48 degree field of view.  
The focal plane for the sensor is a 25 micron pitch, 512 x 2560, 5 micron cutoff HgCdTe array cooled to 80K and operating at a 30 Hz frame rate.  
The sensor instantaneous field of view is 130 (vertical) x 300 (horizontal) microradians.  
Either the 3.7 - 4.25 or 4.6 - 4.85 micron spectral band  is chosen depending on viewing conditions using a warm filter wheel that is part of the optical assembly.  
The large focal plane array (FPA) size is required for detecting dim, sub-pixel targets.

The data processing functions were broken down into two primary subdivisions, the front-end and back-end processing.  
This division was required because of the large amount of data from the sensor that must be handled in real time.  
The primary goal of the front end processing is image processing and the production of candidate detections, which are called exceedances (referred to as "xcds" for brevity in this paper).
The primary goal of the back end processing is the generation and maintenance of tracks, and for the sea skimming missile at the horizon detection mode, correct declaration of threats with a minimum of false alarms. 
This paper focuses on a realizable CS camera and the front end algorithms while the back end processing remains identical.

The detection performance of the system against a surrogate missile target was tested in a series of experiments in September 2005 \cite{LNS06}.  
The sensor was mounted at a height of 65 feet above sea level on the roof of a building 100 meters from the shoreline.  This location provided a clear view to the east over the Atlantic Ocean for the full 45 degree field of view of the sensor.  Data was collected between 0800 and 1500 hours.
The target consisted of a radiometrically calibrated blackbody infrared source constructed using an internally heated metal plate integrated on a towed decoy.  
It provided an in-band signature of approximately 4 watts/ster, which is approximately the expected signal from a subsonic missile.  
The target was towed from a Lear jet flying at an altitude of approximately 2500 feet and speed of 270 knots.  
The target altitude was controlled from the tow aircraft and varied between 50 and 150 feet above the sea surface depending on wind conditions.  
The target can be seen to be above the geometric horizon at ranges exceeding 15 nautical miles for target heights of greater than 50 feet
 Representative sensor imagery is illustrated in Figure  \ref{fig:OrigImage}.  
Along with the towed target there were a variety of small surface vessels, both commercial fishing and pleasure craft, that presented point like sources to the sensor.  

\begin{figure}[tbc]
  \centerline{
   \begin{tabular}{c}
        \includegraphics[scale=1.0]{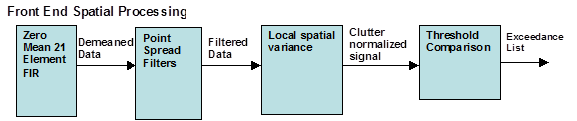}
   \end{tabular}
   }
   \caption{Block diagram for single frame based spatial processing sea skimming missile detection algorithm.}
   \label{fig:FrontEndAlgo}
\end{figure}

Perhaps the main challenge in the missile detection problem is the large number of false alarms (FAs) that tend to arise due to clutter in 
the scene such as sunlight reflecting off the sea.  
The front end processing must minimize these point like sources without losing sight of the target, which is  is relatively dim compared to these bright FAs.
To accomplish this goal, background normalized values are computed by the front end.
Specifically, this two-dimensional spatial image processing consists of:
\begin{enumerate} 
\setlength{\itemsep}{1pt}
  \setlength{\parskip}{0pt}
  \setlength{\parsep}{0pt}
  \item Read in an image frame,
  \item Single frame spatial demeaning (i.e., convolve with a 1-D, 21 element, zero mean filter), 
  \item Four parallel, optical point spread function (PSF) matched filtering, 
  \item Local neighborhood calculation of the spatial variance, 
  \item Computing background normalized values as the filtered results divided by spatial variance, and
  \item Thresholding of the background normalized values in order to identify exceedances (xcds).  
\end{enumerate}
A summary of the original spatial detection algorithm is shown in Figure \ref{fig:FrontEndAlgo}.
For simplicity, this algorithm was slightly modified for the results in this paper by using only one PSF for the matched filtering.
In this paper, applying this algorithm on the collected field data constitutes the conventional approach to missile detection.
It is important to note that each of the above listed steps constitutes a {\it linear} operation on the data.  
This will become important later in the image recovery step of the CS approach.

\begin{figure}[tbc]
  \centerline{
   \begin{tabular}{c}
        \includegraphics[scale=0.3]{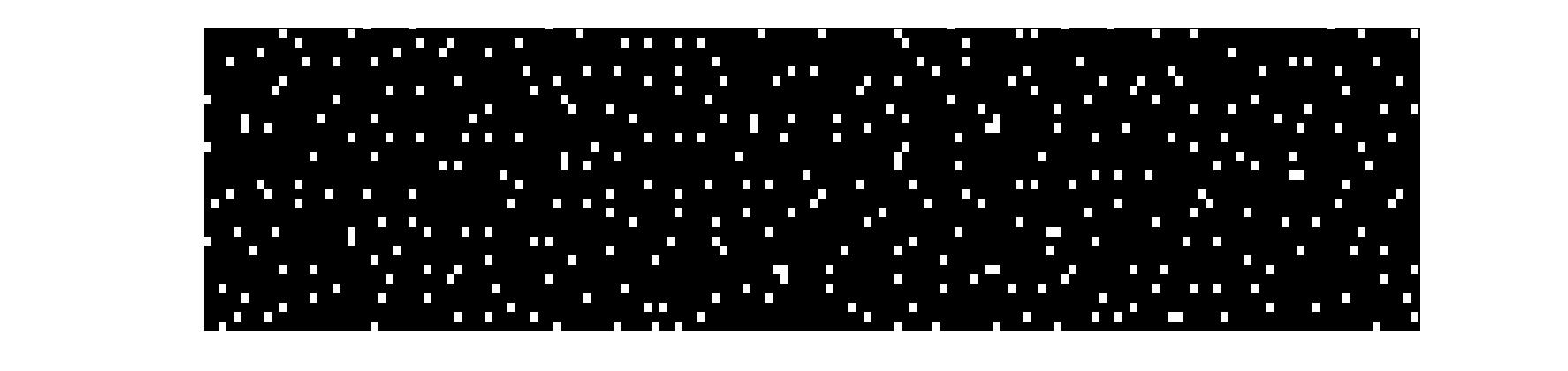}
   \end{tabular}
   }
   \caption{Sample of mask for the 128x640 FPA where 15 out of 16 pixels are blocked (shown in black) and only 1 mask pixel is open (shown in white).}
   \label{mask16}
\end{figure}     

\begin{figure}[tbc]
  \centerline{
   \begin{tabular}{c}
        \includegraphics[scale=0.31]{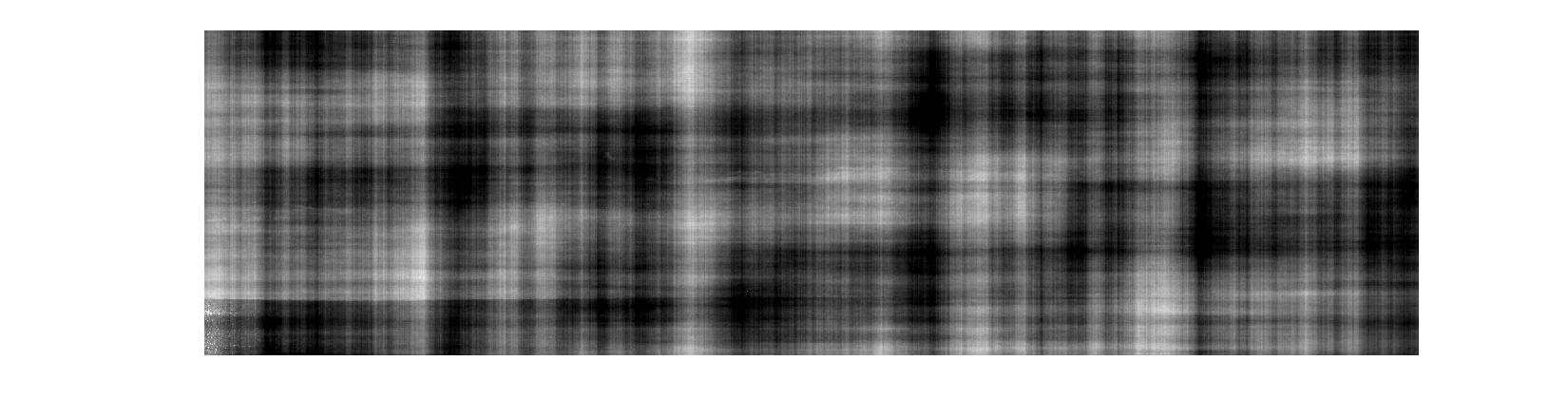}
   \end{tabular}
   }
   \caption{Direct image reconstruction from partial Fourier coefficients where 15 out of 16 pixels are blocked  and only 1 mask pixel is open.}
   \label{PartFourierImage}
\end{figure}

\subsection{CS Missile Detection} 
\label{sec:CSmissile}

The goal of a CS missile detection system is to maintain a high level of correct true target detections while obtaining  only a fraction of the number measurements relative to the number of pixels in the original system.  

Does the sea skimming missile problem fulfill the conditions for finding a practical application of CS?
This particular problem does fit nicely within the CS requirements.
The first requirement for a practical CS application is when applying CS can overcome hardware, financial, or battlefield limitations.
In the missile detection scenario the trend is towards more resolution and smaller pixels in order to better detect dim, sub-pixel missile.
However, a four times increase in number of pixels often translates into an order of magnitude increase in cost.  
Hence, CS can combat this long term trend for more expensive FPAs.

Next, one needs a priori knowledge of what you will be imaging in order to develop the sparsifying and incoherent transforms.
In this situation the objects of interest are only the point sources within the scene.  
Since Fourier transforms are incoherent with delta functions, which are point sources in the discrete realm, they are ideal for this application.
Third is the requirement that the sparsity be orders of magnitude smaller than the number of pixels.
This is true in this case because there is only one missile and only hundreds of point like pixels out of over a million pixels; hence a reduction of four orders of magnitude.
The fourth requirement is incoherence between measurement matrix and the sparsifying transform.
Here the point like objects are sparse in the image domain and as mentioned, the Fourier transforms are ideally incoherent.

The fifth guideline asks ``Is the hardware for the CS sensor realizable?''
Yes; for the envisioned hardware one can use Fourier optics and a mask to measure some fraction of the Fourier coefficients.
For proof of concept and prototyping purposes, we used a simple, random binary mask where each entry is the result of a Bernoulli trial.
As clearly explained in the paper by Lustig, Donoho, and Paula\cite{Lustig07},``Randomness is too important to be left to chance''. 
The choice of the mask is critical to obtaining good results because the mask directly affects the aliasing artifacts.
Hence, several masks of this simple type were tested with our imagery and the best one was used in our experiments in Section \ref{sec:results}.
Two metrics were used to compare masks in order to pick the best; one metric is to count the number of FAs and pick the mask that produced the fewest FAs; the other metric is to compute the Mean Square Error (MSE) of the image reconstructed with the partial Fourier coefficients and pick the mask that produced the smallest error.
The ROC results for both of these masks are displayed in Figure \ref{ROC}, where the "1st mask" is the result of the first metric and the "2nd mask" is the result of the second metric above.

Still, this type of mask is likely quite inefficient and there exists abundance of numerical evidence demonstrating the advantage of variable density sampling strategies \cite{Lustig07, Puy11}, which was not tested in this demonstration.
Furthermore, Romberg has shown \cite{Romberg09} that greater efficiencies are possible with random modulation masks.

For this simulation the scenario consists of a smaller FPA; that is, increase the size of the pixel by 2 in each direction.
However, the mask retains original resolution/pixel size, which in this case is 512x2560.
By creating a mask that blocks out 3 of 4 pixels, each FPA "fat" pixel measures the Fourier coefficient of only the one open mask pixel.
The mask is designed to randomly pick which 3 are blocked and which is open.
Similarly we tested reducing the FPA further -  to only 128x640 and 64x320 pixels and the 512x2560 pixel mask blocks 15 out of 16 or 63 out of 64, respectively.
A sample of a section of the 1 out of 16 mask is shown in Figure \ref{mask16}.

An approach to solving the missile detection problem is to zero fill the missing Fourier coefficient and reconstruct the image with an inverse Fourier transform.  
Figure \ref{PartFourierImage} shows the resulting image from this direct approach of reconstruction from partial Fourier (PF) coefficients where 15 out of 16 pixels are blocked  and only 1 mask pixel is open.
Strong artifacts are clearly visible in the image, which interfere with the processing described above in section \ref{sec:missile}.
However, as shown in the Results section, thresholding the background normalized image for point like objects works reasonably well when the SNR of the target is large.
This implies that the PF coefficients retain information on the point like objects in the frame.

Instead, previous successful examples of compressive sensing in sparse MRI \cite{Lustig07} and radio interferometry \cite{Li11} point to more successful CS approaches for solving the CS missile detection problem.
Following the CLEAN approach\cite{Hogbom74} used in radio interferometry and Donoho, Tsaig, and Starck's stagewise orthogonal matching pursuit \cite{Donoho12},  one can implement a greedy method by iteratively thresholding the background normalized image and removing the brightest points.
In addition, this is equivalent to the nonlinear thresholding scheme described by Lustig, Donoho, and Pauly\cite{Lustig07}.
On the other hand, following the techniques from this same paper on sparse MRI \cite{Lustig07} and a paper describing radio interferometry \cite{Li11} one can analogously solve the following  optimization problem as an $ L_1 $ relaxation approach:
\begin{equation}
\operatorname*{arg\,min}_x \left( \parallel T(x) \parallel_1 + 0.5 \parallel y -  M F x \parallel_2^2    \right)
\end{equation}
where $ x $ is the unknown image, $ y $ is the observed undersampled Fourier coefficients, $ F $ is the Fourier transform, $ M $ is a mask for under-sampling the coefficients, and $ T(x) $ is a function that represents the missile detection algorithm described above in section \ref{sec:missile}.
In other words, the problem constraint has the detection algorithm ``built in''.  
By minimizing $\parallel T(x) \parallel_1$ we are minimizing the number of declared targets.  
Given proper thresholding parameters in $T$, the false alarms will be significantly reduced and only true positives survive the minimization.  
While this is not at all appropriate for pristine image reconstruction, that is not our goal.

There exists many algorithms for solving the  $ L_1 $  optimization problem.
In the Results shown in Section \ref{sec:results} we used the conjugate gradient (CG) method with a total variation prior described in the MRI \cite{Lustig07} 
paper and the robust TwIST method\cite{Bioucas07}.

\begin{table}[tbc]
\caption{Compares results the original algorithm on the original image to CS in various configurations.  The fewer exceedences above the target pixel intensity, the better the algorithm.} 
\label{tab:missile}
\begin{center}       
\begin{tabular}{|l|c|} 
\hline
\rule[-1ex]{0pt}{3.5ex}  \textbf{Image processing} & \textbf{Number of exceedences above target}  \\
\hline
\rule[-1ex]{0pt}{3.5ex}  Original 512x2560 image algorithm & 90   \\
\hline
\rule[-1ex]{0pt}{3.5ex}  CS using 256x1280 PF Coeffs via TwIST or thresholding  & 10   \\
\hline
\rule[-1ex]{0pt}{3.5ex}  CS using 128x640 PF Coeffs via TwIST or thresholding & 23  \\
\hline
\rule[-1ex]{0pt}{3.5ex}  CS using 128x640 PF Coeffs; Remove groups of 5 & 14  \\
\hline
\rule[-1ex]{0pt}{3.5ex}  CS using 64x320 PF Coeffs via TwIST or thresholding & 32  \\
\hline
\rule[-1ex]{0pt}{3.5ex}  CS using 64x320 PF Coeffs; Remove groups of 5 & 8 \\
\hline
\end{tabular}
\end{center}
\end{table}

\subsection{Results} 
\label{sec:results}

This section presents two tests of this CS missile detection system; results from a feasibility test to detect the target in a single frame and a more comprehensive test of missile detection over 500 frames of video.

Initially we performed a first test to determine the feasibility of detecting the surrogate missile target using CS methodology.
This test used a frame where the missile SNR is relatively large. 
Table \ref{tab:missile} compares results the original algorithm on the original image (512x2560) to CS in various configurations.
The original algorithm on the original image is able to find missile target in the background normalized image but, unsurprisingly, there are 90 false alarms (FAs) that are of stronger intensity than the true target pixel.
In this test, the fewer FAs above the target pixel intensity, the better the likelihood of target detection and tracking.

If one doubles the width and height of the pixels, one gets 256x1280, or a quarter of the pixels, which is the case using CS in the next row of Table \ref{tab:missile}.
In our experiments, using either TwIST for L1 optimization solution or simple thresholding, all produced the same results, which was 10 FAs that are of stronger intensity than the true target pixel.
That the CS method improved over the original algorithm with the full image was initially surprising but later tests showed that this was related to the nature of the image frame used in this experiment.
However, this result does provide the proof we were seeking that the concept of CS missile detection is valid.
The next row of Table \ref{tab:missile} gives the results when the resolution of the FPA is reduced once again (to 128x640) and here too the result of 23 brighter xcds is better than the results from the original algorithm.
Test of a non-linear thresholding, where the brightest 5 xcds were iteratively removed as described in Section \ref{sec:CSmissile}, is shown in the next row.  
This demonstrates that the non-linear methodology can improve on simple thresholding.
Finally, the last two rows of the Table show that the CS methods can work for this image with even a lower FPA resolution of 64x320.

Given a satisfactory result from our proof of concept, a more comprehensive test of the CS missile detection system was performed by comparing ROC curves of the target detection over 500 frames of video.
The target signal varies significantly and randomly from frame to frame, offering a challenging test of the CS technique.
The goal is to detect the true target in all 500 frames with a minimum number of FAs.

\begin{figure}[tbc]
  \centerline{
   \begin{tabular}{c}
        \includegraphics[scale=0.5]{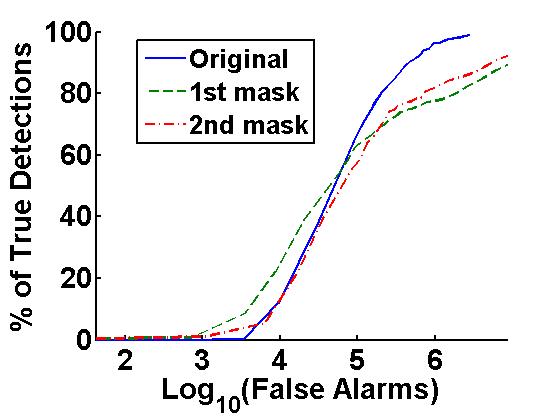}
   \end{tabular}
   }
   \caption{ROC comparing the original missile detection algorithm to a CS algorithm over 500 frames of video containing the surrogate missile target.  The CS method utilized a conjugate gradient method with sparse and TV priors.}
   \label{ROC}
\end{figure}

The conjugate gradient (CG) optimization \cite{Lustig07} weights were both set to 0.0001 to include both the sparsity and total variation (TV) constraints. 
Figure \ref{ROC} shows the ROC curve for the original algorithm and another curve for the CS solution by CG with a total variation (TV) prior at 256x1280 resolution.
Each ROC is computed by processing each frame by its algorithm, applying a set of thresholds, and recording at each threshold the number of FAs and if the true target is detected.
Afterward, the number of FAs and target detections is summed over all of the frames and is used to plot the ROC line.

Figure \ref{ROC} illustrates that our CS method using CG a TV prior at 256x1280 resolution provides a competitive result to the original algorithm.  
Of particular interest is the initial behavior where up until 60\% of the target detections the CS method obtains the target with fewer FAs than the original method.
Of note is that the initial test described above used one of these frames, explaining why in Table  \ref{tab:missile} the CS method obtained fewer FAs than the original algorithm.

Finally we compare computation time for the CS method versus the original algorithm.
Running the original algorithm over 500 frames of video using a Matlab function takes 166 seconds of execution time, which is an order of magnitude slower than real time (at 30 Hz, 500 frames last 16.7 seconds).
This algorithm does run in real time when implemented on an FPGA.
CS using simple thresholding adds a negligible amount of execution time but the performance of the thresholding was unsatisfactory.
Running the CS method using CG with a TV prior at 256x1280 resolution executes in 2300 seconds, which is much greater than real time.
Future work might show that utilizing GPU's and parallel processing techniques will permit running these codes within the real time requirements of these types of systems.

\section{CONCLUSIONS} 

The purpose of this article is to encourage and assist researchers in finding those niche applications where the CS approach provides substantial gain over conventional approaches.
Now that the theory has advanced so significantly, it is time to focus on finding applications where CS will win in real-world comparisons to current conventional state-of-the-art techniques, which include techniques such as super-resolution and image deblurring.
This paper presented lessons learned to help researchers find practical, real-world applications, then used these guidelines to find and show the potential benefit of a new CS application - sea skimming missile detection. 

The results shown here for CS missile detection are only the beginning of the required effort to demonstrate practical target detection.
No effort was taken to optimize the architecture or mask for the missile detection problem, yet competitive results were obtained with this simple CS method using a lower resolution focal plane array.
This is proof of the potential of the CS method for this application and additional research is required on masks, solvers and architectures that certainly will improve on these initial results.

Finally, the primary message is that all of the excitement surrounding CS is necessary and appropriate for encouraging our creativity but we all must take off our "rose colored glasses" and critically judge our ideas, methods and results relative to conventional imaging approaches.

\section*{Acknowledgments} The author gratefully thanks Jonathan M. Nichols and James R. Waterman for their assistance and  the Naval Research Laboratory for supporting this work.


\bibliographystyle{spiebib}   
\bibliography{refs}   


\end{document}